\documentclass[runningheads]{llncs}
\usepackage{amssymb}
\usepackage{amsmath,amssymb,amsfonts}
\usepackage{graphicx}
\usepackage{algorithm,algpseudocode}
\usepackage{tabularx}

\usepackage{caption}
\usepackage{subcaption}
\begin{document}
	%
	%
	%
	\author{Zexuan Yin\inst{1}\orcidID{0000-0002-1306-3858} \and
		Paolo Barucca\inst{1}\orcidID{0000-0003-4588-667X}}
	\authorrunning{Z. Yin et al.}
	%
	\institute{Department of Computer Science, University College London, London, United Kingdom \\
		\email{\{zexuan.yin.20,p.barucca\}@ucl.ac.uk}}
	\title{Stochastic Recurrent Neural Network for Multistep Time Series Forecasting}
	\maketitle
	%
	\begin{abstract}
		Time series forecasting based on deep architectures has been gaining popularity in recent years due to their ability to model complex non-linear temporal dynamics. The recurrent neural network is one such model capable of handling variable-length input and output. In this paper, we leverage recent advances in deep generative models and the concept of state space models to propose a stochastic adaptation of the recurrent neural network for multistep-ahead time series forecasting, which is trained with stochastic gradient variational Bayes. In our model design, the transition function of the recurrent neural network – which determines the evolution of the hidden states – is stochastic rather than deterministic as in a regular recurrent neural network; this is achieved by incorporating a latent random variable into the transition process which captures the stochasticity of the temporal dynamics. Our model preserves the architectural workings of a recurrent neural network for which all relevant information is encapsulated in its hidden states, and this flexibility allows our model to be easily integrated into any deep architecture for sequential modelling. We test our model on a wide range of datasets from finance to healthcare; results show that the stochastic recurrent neural network consistently outperforms its deterministic counterpart.
		
		\keywords{state space models  \and deep generative models \and variational inference.}
	\end{abstract}
	\section{Introduction}
	Time series forecasting is an important task in industry and academia, with applications in fields such as retail demand forecasting \cite{b1}, finance \cite{b2}--\cite{b28}, and traffic flow prediction \cite{b3}. Traditionally, time series forecasting was dominated by linear models such as the autoregressive integrated moving average model (ARIMA), which required prior knowledge about time series structures such as seasonality and trend. With an increasing abundance of data and computational power however, deep learning models have gained much research interest due to their ability to learn complex temporal relationships with a purely data-driven approach; thus requiring minimal human intervention and expertise in the subject matter. In this work, we combine deep learning with state space models (SSM) for sequential modelling. Our work follows recent trend that combines the powerful modelling capabilities of deep learning models with well understood theoretical frameworks such as SSMs.
	
	Recurrent neural networks (RNN) are a popular class of neural networks for sequential modelling. There exists a great abundance of literature on time series modelling with RNNs across different domains \cite{b9}--\cite{b11}. Taking financial time series forecasting as an example, a literature survey \cite{b12} covering papers on the topic between 2005 and 2019 found that 52.5\% of the models involved an RNN component, of which 60.4\% were long short-term memory networks (LSTMs) and 9.89\% were gated recurrent units (GRUs). One limitation of the RNN is that the hidden state transition function is entirely deterministic, which could limit its ability to model processes with high variability \cite{b121}. There is also recent evidence that the performance of RNNs on complex sequential data such as speech and music can be improved when incorporating uncertainty into their hidden states \cite{b13}--\cite{b15}. The authors in \cite{b16}--\cite{b18} incorporated the concept of state space models in their network design to model variability in the transition process of sequential tasks such as video prediction and audio analysis. Inspired by this, we propose and test a stochastic adaptation of the GRU for time series forecasting. The main contributions of our paper are as follows:
	\begin{enumerate}
		\item we propose a novel deep stochastic recurrent architecture for multistep-ahead time series forecasting which leverages the ability of regular RNNs to model long-term dynamics and the stochastic framework of state space models. 
		\item we conduct experiments using publicly available datasets in the fields of finance, traffic flow prediction, air quality forecasting, and disease transmission. Results demonstrate that our stochastic RNN consistently outperforms its deterministic counterpart, and is capable of generating probabilistic forecasts 
	\end{enumerate}
	\section{Related Works}
	\label{sec:rw}
	\subsection{Recurrent Neural Networks}
	The recurrent neural network (RNN) is a deep architecture specifically designed to handle sequential data, and has delivered state-of-the-art performance in areas such as natural language processing \cite{b4}. The structure of the RNN is such that at each time step $t$, the hidden state of the network - which learns a representation of the raw inputs - is updated using the external input for time $t$ as well as network outputs from the previous step $t - 1$. The weights of the network are shared across all time steps and the model is trained using back-propagation. 
	When used to model long sequences of data, the RNN is subject to the vanishing/exploding gradient problem \cite{b5}. Variants of the RNN such as the LSTM \cite{b6} and the GRU \cite{b7} were proposed to address this issue. These variants use gated mechanisms to regulate the flow of information. The GRU is a simplification of the LSTM without a memory cell, which is more computationally efficient to train and offers comparable performance to the LSTM \cite{b8}.   
	\subsection{Stochastic Gradient Variational Bayes}
	The authors in \cite{b15} proposed the idea of combining an RNN with a variational auto-encoder (VAE) to leverage the RNN's ability to capture time dependencies and the VAE's role as a generative model. The proposed structure consists of an encoder that learns a mapping from data to a distribution over latent variables, and a decoder that maps latent representations to data. The model can be efficiently trained with Stochastic Gradient Variational Bayes (SGVB) \cite{b19} and enables efficient, large-scale unsupervised variational learning on sequential data. Consider input $\boldsymbol{x}$ of arbitrary size, we wish to model the data distribution $p(\boldsymbol{x})$ given some unobserved latent variable $\boldsymbol{z}$ (again, of arbitrary dimension). The aim is maximise the marginal likelihood function $p(\boldsymbol{x}) = \int p(\boldsymbol{x}|\boldsymbol{z})p(\boldsymbol{z}) \,d\boldsymbol{z}$, which is often intractable when the likelihood $p(\boldsymbol{x}|\boldsymbol{z})$ is expressed by a neural network with non-linear layers. Instead we apply variational inference and maximise the evidence lower-bound (ELBO):
	\begin{multline} \label{eq1}
		logp(\boldsymbol{x})=log \int p(\boldsymbol{x}|\boldsymbol{z})p(\boldsymbol{z})\,d\boldsymbol{z} = log \int p(\boldsymbol{x}|\boldsymbol{z})p(\boldsymbol{z})\frac{q(\boldsymbol{z})}{q(\boldsymbol{z})}\,d\boldsymbol{z} \\
		\geq \mathbb{E}_{z \sim q(\boldsymbol{z|\boldsymbol{x}})} [logp(\boldsymbol{x}|\boldsymbol{z})] - KL(q(\boldsymbol{z}|\boldsymbol{x})||p(\boldsymbol{z}))= ELBO,		
	\end{multline}
	where $q(\boldsymbol{z}|\boldsymbol{x})$ is the variational approximation to true the posterior distribution $p(\boldsymbol{z}|\boldsymbol{x})$ and $KL$ is the Kullback-Leibler divergence. 
	For the rest of this paper we refer to $p(\boldsymbol{x}|\boldsymbol{z})$ as the decoding distribution and $q(\boldsymbol{z}|\boldsymbol{x})$ as the encoding distribution. The relationship between the marginal likelihood $p(x)$ and the $ELBO$ is given by 
	\begin{multline} \label{eq2}
		logp(\boldsymbol{x})=\mathbb{E}_{z \sim q(\boldsymbol{z|\boldsymbol{x}})} [logp(\boldsymbol{x}|\boldsymbol{z})] - KL(q(\boldsymbol{z}|\boldsymbol{x})||p(\boldsymbol{z})) \\+KL(q(\boldsymbol{z}|\boldsymbol{x})||(p(\boldsymbol{z}|\boldsymbol{x})),
	\end{multline}
	where the third $KL$ term specifies the tightness of the lower bound. The expectation $\mathbb{E}_{z \sim q(\boldsymbol{z|\boldsymbol{x}})} [logp(\boldsymbol{x}|\boldsymbol{z})]$ can be interpreted as an expected negative reconstructed error, and $KL(q(\boldsymbol{z}|\boldsymbol{x})||p(\boldsymbol{z}))$ serves as a regulariser.
	\subsection{State Space Models}
	State space models provide a unified framework for time series modelling; they refer to probabilistic graphical models that describe relationships between observations and the underlying latent variable \cite{b24}. Exact inference is feasible only for hidden Markov models (HMM) and linear Gaussian state space models (LGSS) and both are not suitable for long-term prediction \cite{b20}. SSMs can be viewed a probabilistic extension of RNNs. Inside an RNN, the evolution of the hidden states $\boldsymbol{h}$ is governed by a non-linear transition function $f$: $\boldsymbol{h}_{t+1} = f(\boldsymbol{h}_t, \boldsymbol{x}_{t+1})$ where $\boldsymbol{x}$ is the input vector. For an SSM however, the hidden states are assumed to be random variables. It is therefore intuitive to combine the non-linear gated mechanisms of the RNN with the stochastic transitions of the SSM; this creates a sequential generative model that is more expressive than the RNN and better capable of modelling long-term dynamics than the SSM. There are many recent works that draw connections between SSM and VAE using an RNN. The authors in \cite{b121} and \cite{b13} propose a sequential VAE with nonlinear state transitions in the latent space, in \cite{b21} the authors investigate various inference schemes for variational RNNs, in \cite{b16} the authors propose to stack a stochastic SSM layer on top of a deterministic RNN layer, in \cite{b17} the authors propose a latent transition scheme that is stochastic conditioned on some inferable parameters, the authors in \cite{b22} propose a deep Kalman filter with exogenous inputs, the authors in \cite{b23} propose a stochastic variant of the Bi-LSTM, and in \cite{b26} the authors use an RNN to parameterise a LGSS. 
	\section{Stochastic Recurrent Neural Network}
	\label{sec:srnn}
	\subsection{Problem Statement}
	For a multivariate dataset comprised of $N+1$ time series, the covariates $\boldsymbol{x}_{1:T+\tau} = \{\boldsymbol{x}_1,\boldsymbol{x}_2,... \boldsymbol{x}_{T+\tau}\} \in \mathbb{R}^{N \times (T+\tau)}$ and the target variable $y_{1:T} \in \mathbb{R}^{1 \times T}$. We refer to the period $\{T+1,T+2,... T+\tau\}$ as the prediction period, where $\tau \in \mathbb{Z}^+$ is the number of prediction steps and we wish to model the conditional distribution
	\begin{equation}P(y_{T+1:T+\tau}|y_{1:T}, \boldsymbol{x}_{1:T+\tau}).\label{eq3}\end{equation}
	\subsection{Stochastic GRU Cell}
	Here we introduce the update equations of our stochastic GRU, which forms the backbone of our temporal model:
	\begin{align}
		\boldsymbol{u}_t &= \sigma(\boldsymbol{W}_u\cdot\boldsymbol{x}_t + \boldsymbol{C}_u\cdot\boldsymbol{z}_t +  \boldsymbol{M}_u\cdot\boldsymbol{h}_{t-1} + \boldsymbol{b}_u) \label{eq4}\\
		\boldsymbol{r}_t &= \sigma(\boldsymbol{W}_r\cdot\boldsymbol{x}_t + \boldsymbol{C}_r\cdot\boldsymbol{z}_t + \boldsymbol{M}_r\cdot\boldsymbol{h}_{t-1} + \boldsymbol{b}_r) \label{eq5}\\
		\boldsymbol{\tilde{h}}_t &= tanh(\boldsymbol{W}_h\cdot\boldsymbol{x}_t + \boldsymbol{C}_h\cdot\boldsymbol{z}_t + \boldsymbol{r}_t\odot\boldsymbol{M}_h\cdot\boldsymbol{h}_{t-1} + \boldsymbol{b}_h) \label{eq6}\\
		\boldsymbol{h}_t &= \boldsymbol{u}_t\odot\boldsymbol{h}_{t-1} + (1-\boldsymbol{u}_t)\odot\boldsymbol{\tilde{h}}_t,\label{eq7}	
	\end{align}
	where $\sigma$ is the sigmoid activation function, $\boldsymbol{z}_t$ is a latent random variable which captures the stochasticity of the temporal process, $\boldsymbol{u}_t$ and $\boldsymbol{r}_t$ represent the update and reset gates, $\boldsymbol{W}$, $\boldsymbol{C}$ and $\boldsymbol{M}$ are weight matrices, $\boldsymbol{b}$ is the bias matrix, $\boldsymbol{h}_t$ is the GRU hidden state and $\odot$ is the element-wise Hadamard product. 
	Our stochastic adaptation can be seen as a generalisation of the regular GRU, i.e. when $\boldsymbol{C} = 0$, we have a regular GRU cell \cite{b7}.
	\subsection{Generative Model}
\begin{figure}[ht]
		\begin{subfigure}{0.5\textwidth}
			\centering
			\includegraphics[width=0.8\textwidth]{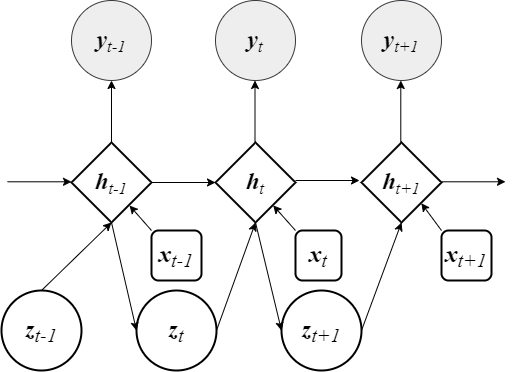}  
			\caption{Generative model}
			\label{fig1}
		\end{subfigure}
		\begin{subfigure}{0.5\textwidth}
			\centering
			\includegraphics[width=0.8\textwidth]{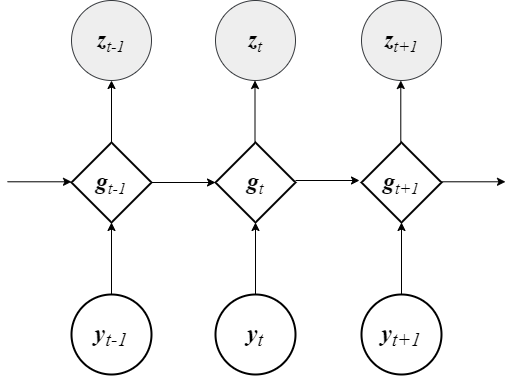}  
			\caption{Inference model}
			\label{fig2}
		\end{subfigure}
		\caption{Proposed generative and inference models}
		\label{fig:fig}
	\end{figure}

	The role of the generative model is to establish probabilistic relationships between the target variable $y_t$, the intermediate variables of interest ($\boldsymbol{h}_t$,$\boldsymbol{z}_t$), and the input $\boldsymbol{x}_t$. Our model uses neural networks to describe the non-linear transition and emission processes, and we preserve the architectural workings of an RNN - relevant information is encoded within the hidden states that evolve with time, and the hidden states contain all necessary information required to estimate the target variable at each time step. A graphical representation of the generative model is shown in Fig \ref{fig1}, the RNN transitions are now stochastic, faciliated by the random variable $\boldsymbol{z}_t$. The joint probability distribution of the generative model can be factorised as follows:
	\begin{equation} \label{eq8}
		p_\theta(y_{2:T},\boldsymbol{z}_{2:T},\boldsymbol{h}_{2:T}|\boldsymbol{x}_{1:T}) =\prod_{t=2}^{T}p_{\theta_1}(y_t|\boldsymbol{h}_t)p_{\theta_2}(\boldsymbol{h}_t|\boldsymbol{h}_{t-1},\boldsymbol{z}_{t},\boldsymbol{x}_t)p_{\theta_3}(\boldsymbol{z}_t|\boldsymbol{h}_{t-1})
	\end{equation}
	where 
	\begin{align}
		p_{\theta_3}(\boldsymbol{z}_t|\boldsymbol{h}_{t-1})&=N(\boldsymbol{\mu}(\boldsymbol{h}_{t-1}),\boldsymbol{\sigma}(\boldsymbol{h}_{t-1})\boldsymbol{\textit{I}})\label{eq9}\\ \boldsymbol{h}_t&=\textit{GRU}(\boldsymbol{h}_{t-1},\boldsymbol{z}_t,\boldsymbol{x}_t)\label{eq10}\\ 
		y_t\sim p_{\theta_1}(y_t|\boldsymbol{h}_t)&=\textit{N}(\mu(\boldsymbol{h}_{t}),\sigma(\boldsymbol{h}_{t})),\label{eq11}
	\end{align}
	where $\textit{GRU}$ is the stochastic GRU update function given by \eqref{eq4}--\eqref{eq7}. \eqref{eq9} defines the prior distribution of $\boldsymbol{z}_t$, which we assume to have an isotropic Gaussian prior (covariance matrix is diagonal) parameterised by a multi-layer perceptron (MLP). When conditioning on past time series for prediction, we use \eqref{eq9}, \eqref{eq10} and the last available hidden state $\boldsymbol{h}_{last}$ to calculate $\boldsymbol{h}_1$ for the next sequence, otherwise we initialise them to $\boldsymbol{0}$. We refer to the collection of parameters of the generative model as $\theta$, i.e. $\theta = \{\theta_1,\theta_2,\theta_3\}$. We refer to \eqref{eq11} as our generative distribution, which is parameterised by an MLP.
	\subsection{Inference Model}
	We wish to maximise the marginal log-likelihood function $logp_\theta(y_{2:T}|\boldsymbol{x}_{2:T})$, however the random variable $\boldsymbol{z}_{t}$ of the non-linear SSM cannot be analytically integrated out. We instead maximise the variational lower bound (ELBO) with respect to the generative model parameters $\theta$ and some inference model parameter which we call $\phi$ \cite{b25}. The variational approximation of the true posterior $p(\boldsymbol{z}_{2:T},\boldsymbol{h}_{2:T}|y_{1:T},\boldsymbol{x}_{1:T})$ can be factorised as follows:
	\begin{equation} \label{eq12}
		q_\phi(\boldsymbol{z}_{2:T},\boldsymbol{h}_{2:T}|y_{1:T},\boldsymbol{x}_{1:T}) = \prod_{t=2}^{T}q_\phi(\boldsymbol{z}_t|y_{1:T})q_\phi(\boldsymbol{h}_t|\boldsymbol{h}_{t-1},\boldsymbol{z}_t,\boldsymbol{x}_t) 
	\end{equation}
	and 
	\begin{equation} \label{eq13}
		q_\phi(\boldsymbol{h}_t|\boldsymbol{h}_{t-1},\boldsymbol{z}_t,\boldsymbol{x}_t) = p_{\theta_2}(\boldsymbol{h}_t|\boldsymbol{h}_{t-1},\boldsymbol{z}_t,\boldsymbol{x}_t),
	\end{equation}
	where $p_{\theta_2}$ is the same as in \eqref{eq8}; this is due to the fact that the GRU transition function is fully deterministic conditioned on knowing $\boldsymbol{z}_t$ and hence $p_{\theta_2}$ is just a delta distribution centered at the GRU output value given by \eqref{eq4}--\eqref{eq7}. The graphical model of the inference network is given in Fig \ref{fig2}. Since the purpose of the inference model is to infer the filtering distribution $q_\phi(\boldsymbol{z}_t|\boldsymbol{y}_{1:t})$, and that an RNN hidden state contains a representation of current and past inputs, we use a second GRU model with hidden states $\boldsymbol{g}_t$ as our inference model, which takes the observed target values $y_t$ and previous hidden state $\boldsymbol{g}_{t-1}$ as inputs and maps $g_t$ to the inferred value of $\boldsymbol{z}_t$:
	\begin{align}
		\boldsymbol{g}_t&=\textit{GRU}(\boldsymbol{g}_{t-1},\boldsymbol{y}_t)\label{eq14}\\
		\boldsymbol{z}_t\sim q_\phi(\boldsymbol{z}_t|\boldsymbol{y}_{1:t})&=N(\boldsymbol{\mu}(\boldsymbol{g}_{t}),\boldsymbol{\sigma}(\boldsymbol{g}_{t})\boldsymbol{\textit{I}}).\label{eq15}
	\end{align} 
	\subsection{Model Training}
	The objective function of our stochastic RNN is the ELBO $\textit{L}(\theta,\phi)$ given by:
	\begin{multline}\label{eq16}
		\textit{L}(\theta,\phi)=\int\int q_\phi log\frac{p_\theta}{q_\phi}d\boldsymbol{z}_{2:T}d\boldsymbol{h}_{2:T}\\ =\sum_{n=2}^{T}\mathbb{E}_{q_\phi} [logp_\theta(y_t|\boldsymbol{h}_t)] - KL(q_\phi(\boldsymbol{z}_t|\boldsymbol{y}_{1:t})||p_\theta(\boldsymbol{z}_t|\boldsymbol{h}_{t-1})), 
	\end{multline}
	where $p_\theta$ and $q_\phi$ are the generative and inference distributions given by \eqref{eq8} and \eqref{eq12} respectively. We seek to optimise the ELBO with respect to decoder parameters $\theta$ and encoder parameters $\phi$ jointly, i.e. we wish to find:
	\begin{equation}\label{eq17}
		(\theta^*,\phi^*)=\operatorname*{argmax}_{\theta,\phi}\textit{L}(\theta,\phi). 
	\end{equation}
	Since we do not backpropagate through a sampling operation, we apply the reparameterisation trick \cite{b19} to write
	\begin{equation}\label{eq18}
		\boldsymbol{z}=\boldsymbol{\mu}+\boldsymbol{\sigma}\odot\boldsymbol{\epsilon}, 
	\end{equation}
	where $\boldsymbol{\epsilon}\sim\boldsymbol{N}(0,\boldsymbol{\textit{I}})$ and we sample from $\boldsymbol{\epsilon}$ instead. The KL divergence term in \eqref{eq16} can be analytically computed since we assume the prior and posterior of $\boldsymbol{z}_t$ to be normally distributed.
	\subsection{Model Prediction}
	Given the last available GRU hidden state $\boldsymbol{h}_{last}$, prediction window $\tau$ and covariates $\boldsymbol{x}_{T+1:T+\tau}$, we generate predicted target values in an autoregressive manner, assuming that at every time step the hidden state of the GRU $\boldsymbol{h}_t$ contains all relevant information up to time $t$. The prediction algorithm of our stochastic GRU is given by Algorithm \ref{alg:prediction}. 
	\begin{algorithm}[H] 
		\caption{Prediction algorithm for stochastic GRU}
		\label{alg:prediction}
		\begin{algorithmic}[1]
			\Require{$\tau,\boldsymbol{h}_{last},\boldsymbol{x}_{T+1:T+\tau}$} 
			\Ensure{$y_{T+1:T+\tau}$}
			\For{$t \gets 1$ to $\tau$}                    
			\State {$\boldsymbol{z}_t$ $\sim$ {$p_{\theta_3}(\boldsymbol{z}_t|\boldsymbol{h}_{last})$}}
			\State {$\boldsymbol{h}_t$ $\gets$ {$\textit{GRU}(\boldsymbol{h}_{last},\boldsymbol{z}_t,\boldsymbol{x}_t)$}}
			\State {$y_t$ $\sim$ {$p_{\theta_1}(y_t|\boldsymbol{h}_{t})$}}
			\State{$\boldsymbol{h}_{last} \gets \boldsymbol{h}_t$}
			\EndFor
		\end{algorithmic}
	\end{algorithm}
	\section{Experiments}
	\label{sec:experiments}
\begin{figure}
		\begin{subfigure}{.5\textwidth}
			\centering
			\includegraphics[width=.8\linewidth]{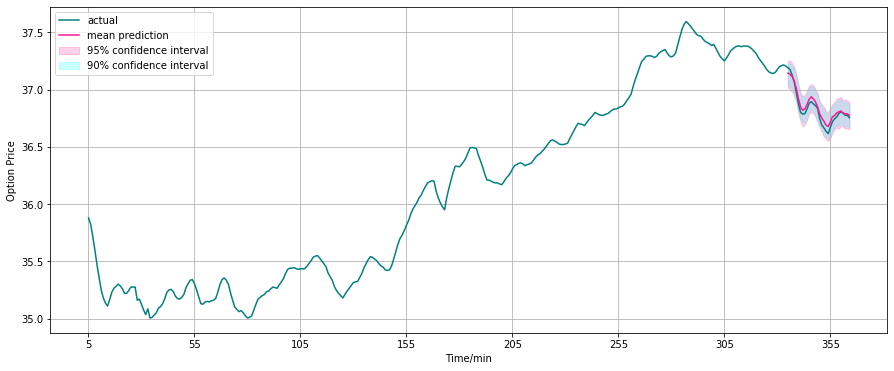}  
			\caption{MSFT call option, strike 190, expiry 17/09/2021}
			\label{fig3}
		\end{subfigure}
		\begin{subfigure}{.5\textwidth}
			\centering
			\includegraphics[width=.8\linewidth]{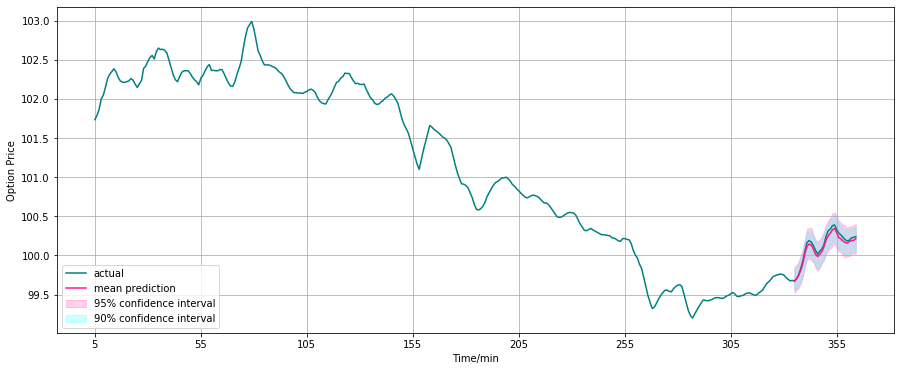}  
			\caption{MSFT put option, strike 315, expiry 16/07/2021}
			\label{fig4}
		\end{subfigure}
		\newline
		\begin{subfigure}{.5\textwidth}
			\centering
			\includegraphics[width=.8\linewidth]{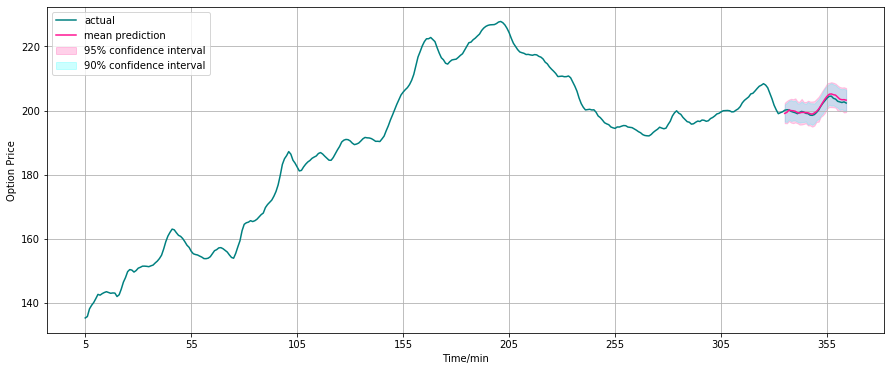}  
			\caption{AMZN put option, strike 3345, expiry 22/01/2021}
			\label{fig5}
		\end{subfigure}
		\begin{subfigure}{.5\textwidth}
			\centering
			\includegraphics[width=.8\linewidth]{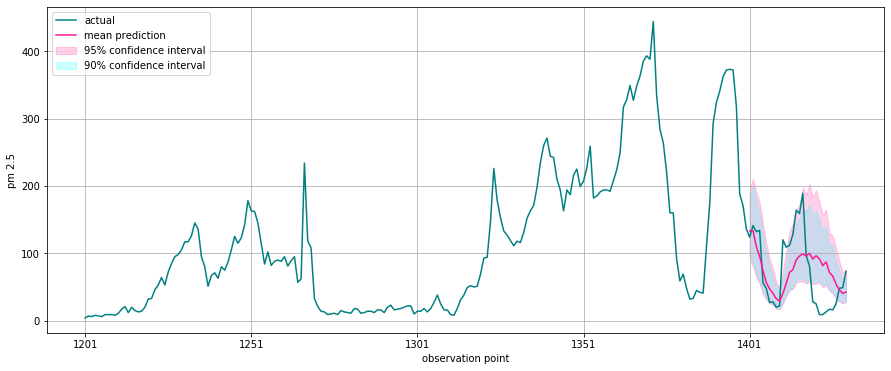}  
			\caption{PM2.5 concentration forecasts up to 30 steps ahead}
			\label{fig6}
		\end{subfigure}
		\newline
		\begin{subfigure}{.5\textwidth}
			\centering
			\includegraphics[width=.8\linewidth]{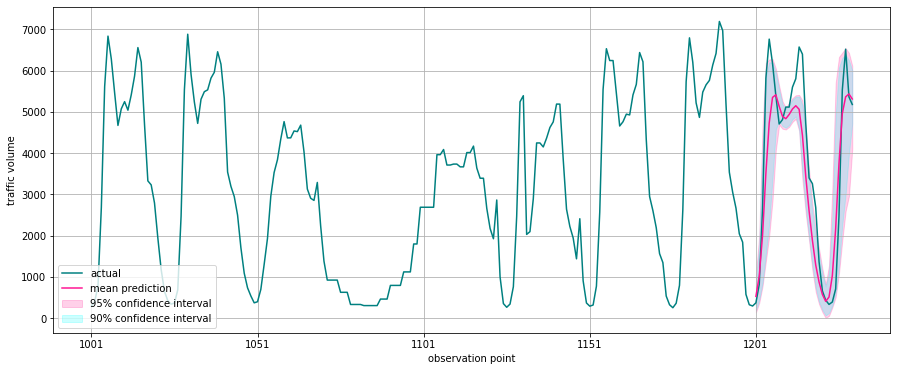}  
			\caption{Traffic volume forecasts up to 30 steps ahead}
			\label{fig7}
		\end{subfigure}
		\begin{subfigure}{.5\textwidth}
			\centering
			\includegraphics[width=.8\linewidth]{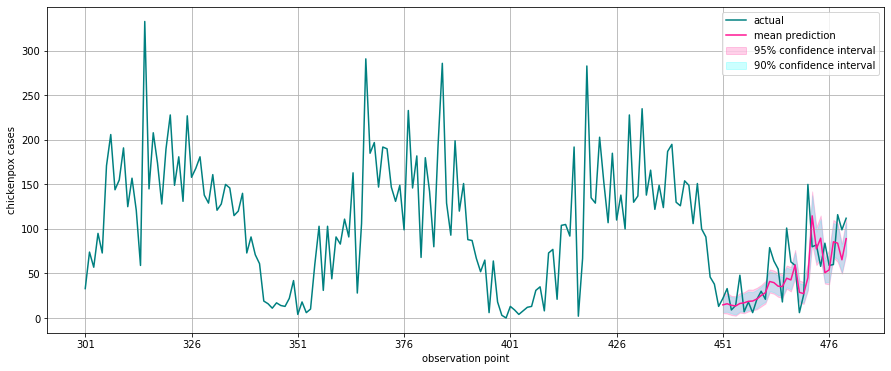}  
			\caption{Hungarian chickenpox cases forecasts up to 30 steps ahead}
			\label{fig8}
		\end{subfigure}
	\end{figure}

	We highlight the model performance on 6 publically available datasets:
	\begin{enumerate}
		\item Equity options trading price time series available from the Chicago Board Options Exchange (CBOE) datashop. This dataset describes the minute-level traded prices of an option throughout the day. We study 3 options with Microsoft and Amazon stocks as underlyings where $\boldsymbol{x}_t=$ underlying stock price and $y_t=$ traded option price
		\item The Beijing PM2.5 multivariate dataset describes hourly PM2.5 (a type of air pollution) concentrations of the US Embassy in Beijing, and is freely available from the UCI Machine Learning Repository. The covariates we use are $\boldsymbol{x}_t=$ temperature, pressure, cumulated wind speed, Dew point, cumulated hours of rainfall and cumulated hours of snow, and $y_t=$ PM2.5 concentration. We use data from 01/11/2014 onwards
		\item The Metro Interstate Traffic Volume dataset describes the hourly interstate 94 Westbound traffic volume for MN DoT ATR station 301, roughly midway between Minneapolis and ST Paul, MN. This dataset is available on the UCI Machine Learning Repository. The covariates we use in this experiment are $\boldsymbol{x}_t=$ temperature, mm of rainfall in the hour, mm of snow in the hour, and percentage of cloud cover, and $y_t=$ hourly traffic volume. We use data from 02/10/2012 9AM onwards
		\item The Hungarian Chickenpox dataset describes weekly chickenpox cases (childhood disease) in different Hungarian counties. This dataset is also available on the UCI Machine Learning Repository. For this experiment, $y_t=$ number of chickenpox cases in the Hungarian capital city Budapest, $\boldsymbol{x}_t=$ number of chickenpox cases in Pest, Bacs, Komarom and Heves, which are 4 nearby counties. We use data from 03/01/2005 onwards
	\end{enumerate}
	We generate probabilistic forecasts using 500 Monte-Carlo simulations and we take the mean predictions as our point forecasts to compute the error metrics. We provide graphical illustrations of the prediction results in Fig \ref{fig3}--\ref{fig8}. We compare our model performance against an AR(1) model assuming the prediction is the same as the last observed value ($y_{T+\tau}=y_T$), and a standard LSTM model. For the performance metric, we normalise the root-mean-squared-error (rmse) to enable comparison between time series:
	\begin{equation} \label{eq19}	
		nrmse = \frac{\sqrt{\frac{\sum_{i=1}^{N}(y_i-\hat{y}_i)^2}{N}}}{\bar{y}},	
	\end{equation}
	where $\bar{y} = mean(y)$, $\hat{y}_i$ is the mean predicted value of $y_i$, and N is the prediction size. For replication purposes, in Table \ref{tab6} we provide (in order): number of training, validation and conditioning steps, (non-overlapping) sequence lengths used for training, number of prediction steps, dimensions of $\boldsymbol{z}_t$, $\boldsymbol{h}_t$ and $\boldsymbol{g}_t$, details about the MLPs corresponding to \eqref{eq9}($\boldsymbol{z}_t$ prior) and \eqref{eq11}($\boldsymbol{z}_t$ post) in the form of (n layers, n hidden units per layer), and lastly the size of the hidden states of the benchmark LSTM.
	we use the ADAM optimiser with a learning rate of 0.001. 
	\begin{table}[h]
		\centering
		\caption{Model and training parameters}
		\label{table1}
		\resizebox{\textwidth}{!}{
			\begin{tabular}{|l|l|l|l|l|l|l|l|l|l|l|l|}
				\hline
				dataset & train & val & cond & seq length & pred & $\boldsymbol{z}_t$ & $\boldsymbol{h}_t$ & $\boldsymbol{g}_t$ & $\boldsymbol{z}_t$ prior & $\boldsymbol{z}_t$ post & LSTM hid\\
				\hline
				Options & 300 & 30 & 10 & 10 & 30 & 50 & 64 & 64 & (4,64) & (4,64) & 64\\
				\hline
				PM2.5 & 1200 & 200 & 10 & 10 & 30 & 50 & 64 & 64 & (4,64) & (4,64) & 64\\
				\hline
				Traffic volume & 1000 & 200 & 20 & 20 & 30 & 30 & 128 & 128 & (4,128) & (4,128) & 128\\
				\hline
				Hungarian chickenpox & 300 & 150 & 10 & 10 & 30 & 50 & 128 & 128 &(4,128) & (4,128) & 128\\
				\hline
		\end{tabular}}
		\label{tab6}
	\end{table}
	\begin{table}[h]
		\centering
		\caption{nrmse for 30 steps-ahead options price predictions}
		\label{table1}
		\begin{tabular}{|l|l|l|l|l|}
			\hline
			Option& 
			Description& 
			Ours & AR(1)& LSTM \\
			\hline
			MSFT call& 
			strike 190, expiry 17/09/2021& 
			\textbf{0.0010}&0.0109&0.0015 \\
			\hline
			MSFT put&strike 315, expiry 16/07/2021&\textbf{0.0004}&0.0049&0.0006\\		
			\hline
			AMZN put&strike 3345, expiry 22/01/2021&\textbf{0.0032}&0.0120&0.0038\\
			\hline
		\end{tabular}
		\label{tab1}
		\centering
		\caption{nrmse for 30 steps-ahead PM2.5 concentration predictions}
		\label{table2}
		\begin{tabular}{|l|l|l|l|l|l|l|}
			\hline
			steps& 
			5&10&15&20&25&30\\
			\hline
			Ours&\textbf{0.1879}&\textbf{0.2474}&\textbf{0.4238}&\textbf{0.4588}&\textbf{0.6373}&\textbf{0.6523}\\
			\hline
			AR(1)&0.3092&1.0957&0.7330&0.6846&1.0045&1.1289\\
			\hline
			LSTM&0.4797&0.6579&0.4728&0.4638&0.8324&0.8318\\
			\hline
		\end{tabular}
		\label{tab2}
		\centering
		\caption{nrmse for 30 steps-ahead traffic volume predictions}
		\label{table3}
		\begin{tabular}{|l|l|l|l|l|l|l|}
			\hline
			steps& 
			5&10&15&20&25&30\\
			\hline
			Ours&\textbf{0.4284}&\textbf{0.2444}&\textbf{0.2262}&\textbf{0.2508}&\textbf{0.2867}&\textbf{0.2605}\\
			\hline
			AR(1)&1.2039&1.0541&1.0194&1.0283&1.1179&1.0910\\
			\hline
			LSTM&0.8649&0.5936&0.4416&0.4362&0.5591&0.5446\\
			\hline
		\end{tabular}
		\label{tab3}
		\centering
		\caption{nrmse for 30 steps-ahead Hungarian chickenpox predictions}
		\label{table4}
		\begin{tabular}{|l|l|l|l|l|l|l|}
			\hline
			steps& 
			5&10&15&20&25&30\\
			\hline
			Ours&\textbf{0.6585}&\textbf{0.6213}&\textbf{0.5795}&\textbf{0.5905}&\textbf{0.6548}&\textbf{0.5541}\\
			\hline
			AR(1)&0.7366&0.7108&0.9126&0.9809&1.0494&1.0315\\
			\hline
			LSTM&0.7215&0.6687&0.9057&1.0717&0.8471&0.7757\\
			\hline
		\end{tabular}
		\label{tab4}
		\centering
		\caption{nrmse of MLP benchmark and our proposed model for 30 steps-ahead forecasts}
		\label{table5}
		\begin{tabular}{|l|l|l|l|l|l|l|}
			\hline
			& 
			MSFT call&MSFT put&AMZN put&PM2.5&Metro&\hspace{0pt}Chickenpox\\
			\hline
			Ours&\textbf{0.0010}&\textbf{0.0004}&\textbf{0.0032}&\textbf{0.6523}&\textbf{0.2605}&\textbf{0.5541}\\
			\hline
			MLP&0.0024&0.0005&0.0141&0.7058&0.6059&0.5746\\
			\hline
		\end{tabular}
	\end{table}
	In Table \ref{tab1}--\ref{table4} we observe that the nrmse of the stochastic GRU is lower than its deterministic counterpart for all datasets investigated and across all prediction steps. This shows that our proposed method can better capture both long and short-term dynamics of the time series. With respect to multistep time series forecasting, it is often difficult to accurately model the long-term dynamics. Our approach provides an additional degree of freedom facilitated by the latent random variable which needs to be inferred using the inference network; we believe this allows the stochastic GRU to better capture the stochasticity of the time series at every time step. In Fig \ref{fig7} for example, we observe that our model captures well the long-term cyclicity of the traffic volume, and in Fig \ref{fig6} where the time series is much more erratic, our model can still accurately predict the general shape of the time series in the prediction period. 
	
	To investigate the effectiveness of our temporal model, we compare our prediction errors against a model without a temporal component, which is constructed using a 3-layer MLP with 5 hidden nodes and ReLU activation functions. Since we are using covariates in the prediction period \eqref{eq3}, we would like to verify that our model can outperform a simple regression-type benchmark which approximates a function of the form $y_t=f_\psi(\boldsymbol{x_t})$; we use the MLP to parameterise the function $f_\psi$. We observe in Table \ref{table5} that our proposed model outperforms a regression-type benchmark for all the experiments, which shows the effectiveness of our temporal model. It is also worth noting that in our experiments we use the actual values of the future covariates. In a real forecasting setting, the future covariates themselves could be outputs of other mathematical models, or they could be estimated using expert judgement. 
	\section{Conclusion}
	In this paper we have presented a stochastic adaptation of the Gated Recurrent Unit which is trained with stochastic gradient variational Bayes. Our model design preserves the architectural workings of an RNN, which encapsulates all relevant information into the hidden state, however our adaptation takes inspiration from the stochastic transition functions of state space models by injecting a latent random variable into the update functions of the GRU, which allows the GRU to be more expressive at modelling highly variable transition dynamics compared to a regular RNN with deterministic transition functions. We have tested the performance of our model on different publicly available datasets and results demonstrate the effectiveness of our design. Given that GRUs are now popular building blocks for much more complex deep architectures, we believe that our stochastic GRU could prove useful as an improved component which can be integrated into sophisticated deep learning models for sequential modelling.  
	\paragraph*{Acknowlegements}
	We would like to thank Dr Fabio Caccioli (Dpt of Computer Science, UCL) for proofreading this manuscript and for his questions and feedback.

\end{document}